%% file: neurips_2026.tex
\documentclass{article}

\usepackage[main,final]{neurips_2026}

\usepackage[utf8]{inputenc} 
\usepackage[T1]{fontenc}    
\usepackage{hyperref}       
\usepackage{url}            
\usepackage{booktabs}       
\usepackage{amsfonts}       
\usepackage{nicefrac}       
\usepackage{microtype}      
\usepackage{xcolor}         
\usepackage{multirow}
\usepackage{multicol}
\usepackage{graphicx}
\usepackage{colortbl}
\usepackage{arydshln}
\usepackage{amsmath}
\usepackage{caption}       
\usepackage{fancyvrb}      
\usepackage{tcolorbox}     
\tcbuselibrary{skins, breakable}
\usepackage{fvextra} 
\usepackage{algorithm}
\usepackage{algpseudocode}
\usepackage{wrapfig}

\definecolor{swecream}{RGB}{255,247,236}

\newcommand{\pub}[1]{{\color{gray}{\tiny{[{#1}]\!}}}}

\title{\textsc{LensDesigner}: A Self-Improving Agent for \\ Optical Lens Design}

\author{%
  \textbf{Lei Sun}$^{1,*}$\quad
  \textbf{Haoran Liang}$^{1,*}$\quad
  \textbf{Dannong Xu}$^{1}$\quad
  \textbf{Yao Gao}$^{2}$\quad
  \textbf{Yuyu Geng}$^{2}$ \\
  \textbf{Jinjin Gu}$^{1}$\quad
  \textbf{Kaiwei Wang}$^{2,\dagger}$\quad
  \textbf{Danda Pani Paudel}$^{1}$\quad
  \textbf{Luc Van Gool}$^{1}$ \\
  $^{1}$INSAIT, Sofia University ``St.\ Kliment Ohridski'' \\
  $^{2}$Zhejiang University \\
  $^{*}$Equal contribution.\quad $^{\dagger}$Corresponding author. \\
  \href{mailto:wangkaiwei@zju.edu.cn}{\texttt{leo\_sun@zju.edu.cn}}\quad
  \href{mailto:leo_sun@zju.edu.cn}{\texttt{wangkaiwei@zju.edu.cn}}
}

\begin{document}

\maketitle

\begin{abstract}
Optical lens design is a complex, non-convex optimization challenge that relies heavily on human experience and intuition. Existing optimized-based automatic lens design methods struggle to navigate this vast parameter space without meticulous manual tuning. In this paper, we present \textsc{LensDesigner}, an autonomous agent framework that mirrors the problem-solving workflow of expert opticians. To overcome the initial cold start problem, we construct LensLib100K, an extensive optical lens library, and employ Optics-Aware Retrieval to supply physically valid structural seeds. Within an interactive physical simulation environment, the agent executes macroscopic orchestration while receiving immediate optical feedback. Furthermore, we introduce a continuous self-evolving mechanism guided by a curriculum agent. By iteratively solving design tasks with progressively increasing difficulty, the agent autonomously extracts, accumulates, and reuses design heuristics, effectively evolving its optical lens design expertise over time. At the evaluation level, we introduce LensArena, a standardized evaluation benchmark comprising $120$ diverse optical design tasks, covering extreme configurations. Extensive experiments on this benchmark demonstrate that \textsc{LensDesigner} significantly outperforms publicly available baseline algorithms, achieving superior success rates and optimization efficiency. We hope this work sheds light on the emerging field of intelligent optics. The code will be publicly available.
\end{abstract}

\input{color_box_define}

\section{Introduction}
\label{sec:intro}
\begin{figure}[t]
    \centering
    \includegraphics[width=0.99\linewidth]{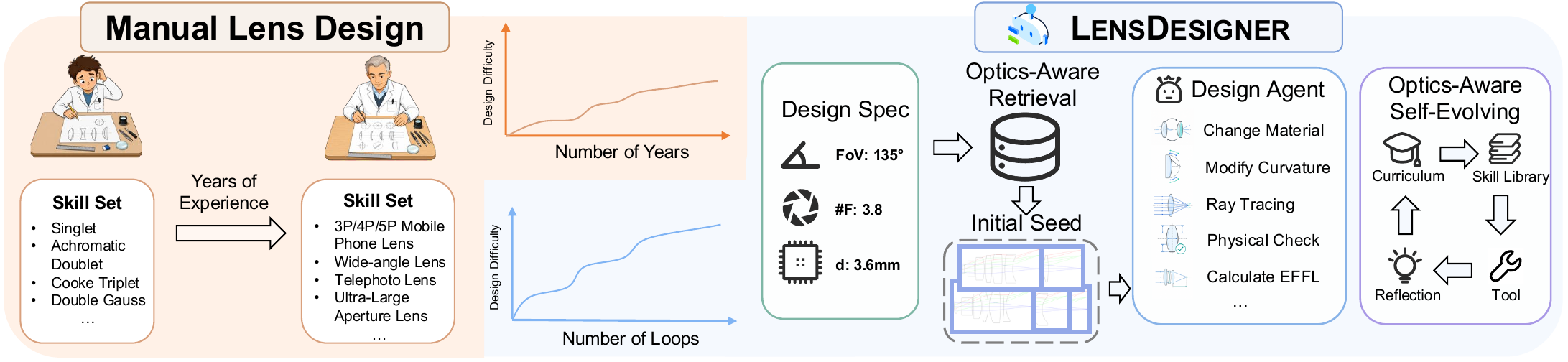}
    \caption{Manual lens design is a highly experience-dependent task that typically requires years of accumulated human expertise. Similarly, \textsc{LensDesigner} mimics this trajectory through a self-evolving mechanism, continuously accumulating skills to advance its optical design capabilities.}
    \vspace{-20pt}
    \label{fig:teaser}
\end{figure}

Optical lens design is a fundamental pillar of modern science and technology, underpinning advancements in diverse fields ranging from consumer electronics and augmented reality to medical imaging and space exploration. However, designing a high-quality lens system is a notoriously challenging, non-convex optimization problem involving a vast, high-dimensional parameter space (e.g., curvature, thickness, glass materials). Crucially, navigating this complex space is intrinsically a highly \textit{experience-dependent} endeavor and heavily relies on human expert intuition. As show in Fig.~\ref{fig:teaser}, a typical design process requires an experienced optician to propose an initial lens structure based on decades of accumulated domain knowledge, trial, and error, followed by iterative refinements using commercial optical simulation software like \textit{Zemax}~\cite{zemax} or \textit{CODE V}. Without this experience-driven high-quality initial starting point, traditional optimization algorithms easily fall into local optima, making the process highly inefficient.

Recently, artificial intelligence has shown great potential in accelerating scientific discovery. However, applying AI to optical design remains fundamentally bottlenecked by the limitations of current methodologies. For instance, approaches utilizing traditional evolutionary algorithms for lens optimization ~\cite{gao2025exploring,gao2025neuro,yang2024curriculum} demand extensive fine-grained parameter tuning that is difficult for users lacking expert intuition. Moreover, these techniques suffer from prohibitive computational costs, frequently needing several days to successfully converge on a single lens design. Conversely, methods relying on simple database query and matching~\cite{cote2021deep} fail to handle novel target specifications outside their existing scope, lacking the autonomous intelligence needed for original design. 

Meanwhile, although Large Language Models (LLMs) are not inherently designed for precise numerical calculation or optimization, they encode rich optical domain knowledge. More importantly, recent advancements in autonomous agents have proven that LLMs can effectively utilize memory and skill modules to continuously learn, and self-evolve~\cite{wang2023voyager,shinn2023reflexion,xia2025agent0unleashingselfevolvingagents,xia2026skillrlevolvingagentsrecursive,acikgoz2026toolr0selfevolvingllmagents}. Therefore, while LLM agents possess immense potential to tackle such highly experience-dependent tasks, their application in automated optical lens design remains an unexplored frontier.

In this paper, we propose \textsc{LensDesigner}, an autonomous agent framework that mirrors the workflow of human experts. Experienced opticians rarely design from scratch. Instead, they rely on known configurations as initial seeds. We replicate this crucial heuristic by constructing a large-scale optical lens library, dubbed LensLib100K, based on an enhanced implementation of the state-of-the-art optimization algorithm QGSO~\cite{gao2025exploring}. This comprehensive dataset encompasses a wide spectrum of common lens specifications. Leveraging this library, we design Optics-Aware Retrieval strictly grounded in fundamental optical design principles. For every new design requirement, this module retrieves the most suitable configuration from the library to serve as a physically valid initial seed.

Once the initial seeds are retrieved, \textsc{LensDesigner} undertakes substantial macro orchestration within an interactive optical design environment. The agent autonomously executes actions, such as inserting new elements or altering glass materials, and receives immediate evaluation feedback from an underlying physical simulator regarding the optical merit function and structural validity. 

To simulate the experience accumulation of a human optician, our framework introduces a continuous evolution mechanism driven by a dedicated curriculum agent. This curriculum agent continuously generates physically valid optical design tasks with progressively increasing difficulty. As \textsc{LensDesigner} attempts to solve these generated problems, it actively interacts with the environment and reflects on both successful and erroneous design trajectories based on the simulator feedback. By analyzing these outcomes, the LLM extracts generalized optical design heuristics and commits these validated strategies to an optical skill library for future reuse. This continuous cycle of problem solving, reflection, and memorization enables \textsc{LensDesigner} to continuously evolve, transforming its approach from basic trial and error into mature design intuition. Finally, the framework invokes traditional local optimization algorithms to perform meticulous parameter tuning, guiding the newly synthesized lens to its precise local optimum.

Beyond these methodological advancements, the field of automated optical design currently lacks a standardized benchmark to fairly evaluate the capabilities of different algorithms. Hence, we introduce LensArena, a comprehensive benchmark comprising 120 diverse optical design tasks. This challenging benchmark systematically covers both common lens configurations and extreme optical systems, including ultra-wide-angle lenses with a Field of View (FoV) up to 180° and large-aperture lenses with an F-number (\#F) as small as 0.7, etc. Utilizing LensArena, we rigorously test all publicly available automatic lens design algorithms to establish a comprehensive evaluation standard. Extensive experiments on this benchmark demonstrate that our proposed \textsc{LensDesigner} significantly outperforms existing methods by consistently generating valid lens structures, achieving the highest success rate, and yielding the smallest (root-mean-square) RMS spot radius for optimal imaging quality.

In summary, we deliver the following contributions:
\begin{itemize}
    \item We propose \textsc{LensDesigner}, a novel agent framework specifically designed to automate the highly empirical and complex process of optical lens design by mirroring human expert intuition.

    \item We introduce a continuous evolution mechanism driven by a curriculum agent, enabling the system to autonomously extract, accumulate, and reuse optical design heuristics, progressively advancing its lens design expertise through iterative problem solving.

    \item We construct LensLib100K, an extensive optical lens library, and introduce LensArena, a standardized evaluation benchmark, laying a robust foundation for future intelligent optical design research.
\end{itemize}

\section{Related Work}
\label{sec:related_work}

\paragraph{Automatic Lens Design.}
Traditional optical design is highly empirical, relying on expert intuition and software tools like Zemax that function as numerical calculators rather than autonomous decision-makers. Existing automation attempts include global optimization algorithms~\cite{gao2025exploring, liu2025global}, neural networks~\cite{yang2024curriculum}, and retrieval-based methods like LensNet~\cite{cote2021deep}. However, these approaches often require meticulous manual tuning, suffer from prolonged convergence times, or are constrained by fixed database distributions. In contrast, \textsc{LensDesigner} enables rapid, text-driven generation of compliant configurations, bypassing the need for specialized domain expertise and intensive parameter adjustment.

\paragraph{LLM-based Agents for Scientific Design.}
Large Language Models (LLMs) such as GPT-5 and Qwen-3 have evolved into sophisticated reasoning engines capable of powering autonomous agents~\cite{singh2025openai, yang2025qwen3}. By integrating tool utilization and environmental feedback, these agents have revolutionized specialized fields including chemistry~\cite{bran2023chemcrow}, robotics~\cite{openclaw2026}, and software engineering~\cite{zhang2024autocoderover}. Despite these successes, the application of autonomous entities to the highly empirical and numerically sensitive domain of optical design remains largely unexplored. We bridge this gap by introducing a self-evolving agentic framework specifically optimized for the physics-heavy requirements of lens design. For more detailed related work, please kindly refer to Appendix~\ref{supp:related_work}.

\section{\textsc{LensDesigner}}
\label{sec:method}
In this work, we focus on classical imaging lens design, whose goal is to arrange and optimize the shape, spacing, and materials of refractive components to minimize aberrations and project a sharp image onto a sensor. This optimization process ensures the system meets strict imaging specifications, such as field-of-view (FoV), F-number (\#F), and effective focal length (EFFL). As a pioneering effort in this domain, our work focuses exclusively on spherical lenses.

\subsection{Lens Design Preliminaries}
\label{sec:pre}
A standard system comprises four fundamental structures: the object plane originating the target scene, the optical elements sequentially refracting incoming light, the stops restricting light bundles to control brightness, and the image plane capturing the final picture. 

In a lens system, the Field of View (FoV) defines the maximum angular extent of the observable environment. The F number (\#F) characterizes the size of the optical aperture, representing the ratio of the focal length to the entrance pupil diameter and dictating the overall light-gathering capability. The optical format ($d$) denotes the physical diagonal diameter of the image sensor, while its corresponding semi-image height $y$ defines the maximum radial boundary of focused light. Furthermore, the Effective Focal Length (EFFL) represents the focal length of the optical system, and the Total Track (TOTR) indicates the overall physical length of the entire lens assembly. These configurations fundamentally determine the physical characteristics and practical applications of the lens. Finally, the Root Mean Square (RMS) of the focal spot radius serves as the most fundamental quantitative indicator for reflecting the actual imaging sharpness and evaluating overall optical aberrations. Further theoretical details are provided in Appendix~\ref{supp:preliminaries}.

\subsection{Overview}
\label{sec:overview}
\begin{figure}
    \centering
    \includegraphics[width=0.95\linewidth]{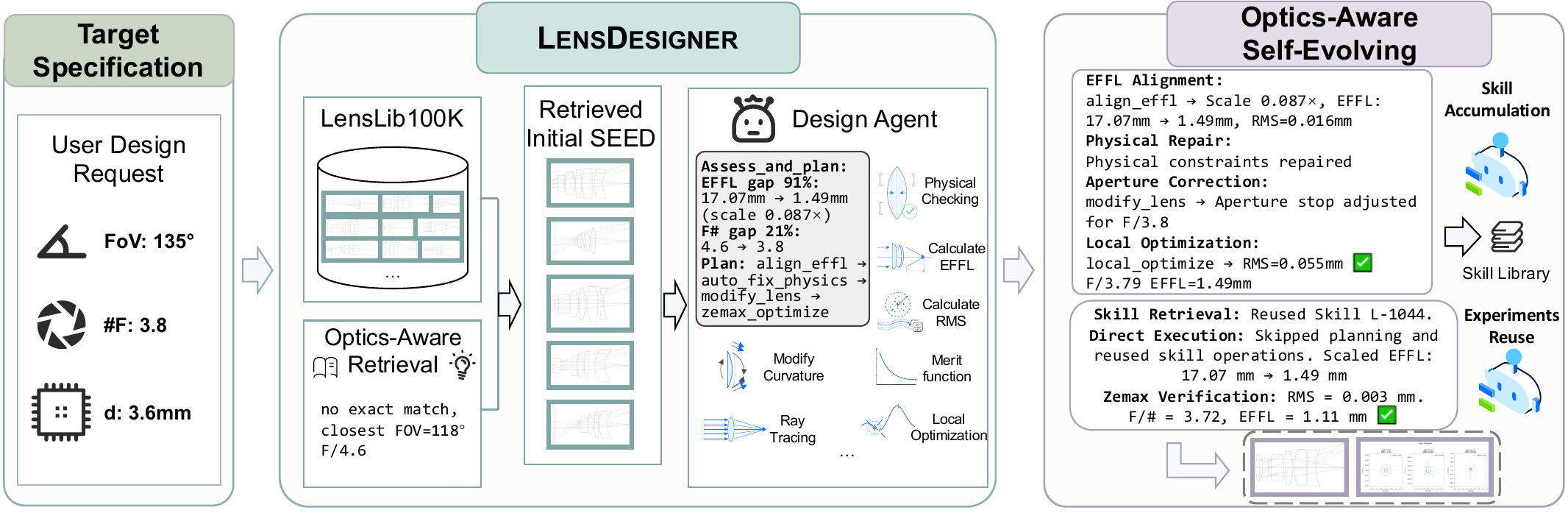}
    \vspace{-5pt}
    \caption{Main pipeline of \textsc{LensDesigner}. Given a design query, the framework retrieves the closest seeds from {LensLib100K}, followed by Macro Orchestration for systematic optimization and Optics-Aware Self-Evolving for continuous capability enhancement via a curriculum-based agent.}    
    \label{fig:pipeline}
    \vspace{-8pt}
\end{figure}

Inspired by the empirical workflow of human optical engineers, \textsc{LensDesigner} operates as a closed-loop autonomous system that integrates knowledge retrieval, interactive reasoning, and self-evolution. The overall pipeline is illustrated in Fig.~\ref{fig:pipeline}.

The Optics-Aware Retrieval identifies physically compatible initial seeds from our LensLib100K to ensure a good starting point. Then, the framework utilizes Macro Orchestration to coordinate an Optical Tool Suite, which executes structural modifications and local optimizations under strict physical guardrails. By continuously reflecting on and accumulating expertise throughout the design process, the system achieves a closed-loop evolution of design intuition.

\subsection{LensLib100K}
\label{sec:lenslib}

\paragraph{Large-Scale Lens Generation.}~Considering the highly free and empirical process of optical design, designing a large number of lenses manually is practically infeasible. Although current automated optical design methods remain imperfect, they can produce a substantial volume of satisfactory designs under simpler conditions, such as narrow FoV and large \#F, provided they are guided by manual parameter tuning over several months.
QGSO utilizes a genetic algorithm-based framework \cite{katoch2021review} to optimize a population of randomly initialized lens structures through iterative mutation and selection. By integrating hybrid global and local optimization to produce diverse, high-quality designs, it currently represents the state-of-the-art in optimization-based lens design.

Building upon the QGSO framework, we introduce two enhancements to bridge the gap between simulation-based optimization and practical optical engineering: First, we enforce the use of real glass catalogs instead of virtual materials, ensuring the optimized parameters are practically manufacturable. Second, we integrate a ray aiming~\cite{cote2023differentiable} mechanism to guarantee high-fidelity ray tracing under large FoVs. Overall, these improvements enhance the manufacturability of the design outcomes, which makes the produced lens more applicable.

\begin{wrapfigure}{l}{0.5\textwidth}
    \centering
    \includegraphics[width=\linewidth]{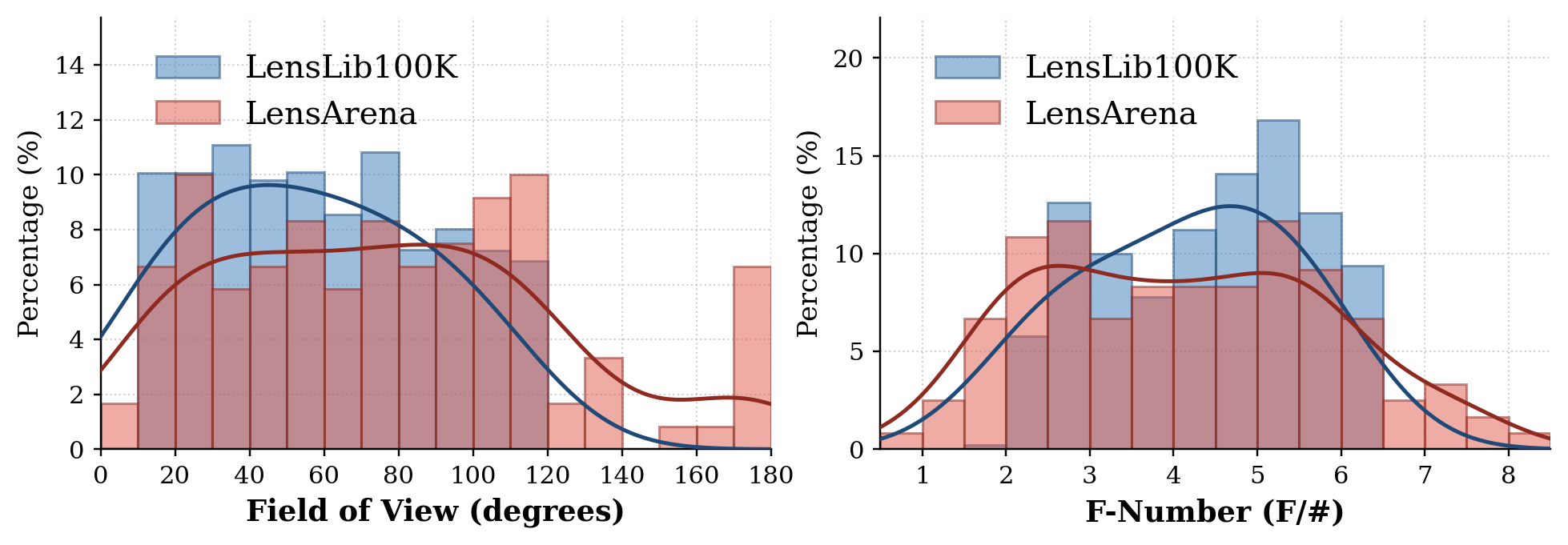}
    \vspace{-5pt}
    \caption{Distribution of LensLib100K and LensArena. LensArena spans more challenging edge cases.}
    \label{fig:lenslib}
    \vspace{-8pt}
\end{wrapfigure}

\paragraph{Data Distribution}~Leveraging our extended QGSO framework, we invested approximately $2000$ GPU hours of intensive computation to generate a comprehensive collection of $180000$ distinct lens designs. These generated configurations exhibit remarkable diversity across critical specifications. Specifically, the FoV spans from $8^\circ$ to $118^\circ$, and the \#F ranges from $1.6$ to $6.0$. Data distribution is shown in Fig.~\ref{fig:lenslib}. To the best of our knowledge, our LensLib100K represents the most extensive optical lens database currently available to the research community. It not only serves as a practical foundation for direct structural retrieval within our \textsc{LensDesigner} framework but also offers a substantial collection of high-quality training data to support future research in data-driven automatic optical design.

\paragraph{Optics-Aware Retrieval}
To effectively leverage LensLib100K, we introduce an Optics-Aware Retrieval mechanism acting as a physics-grounded selector. Retrieval criteria used in conventional semantic RAG methods are fundamentally ill-suited for optical design. In optics, a system's physical properties are determined by the complex interplay of multiple query requirements, rather than isolated parameters. Consequently, retrieving candidates based merely on the closest individual numerical values does not necessarily yield the optimal solution. 
To overcome this, we utilize simulator-in-the-loop labeling, where every lens undergoes offline paraxial ray tracing to pre-compute physical metrics before indexing.
This ensures retrieval relies purely on deterministic optical performance. Furthermore, we score candidates using a novel asymmetric optical prior distance. This metric heavily penalizes deviations in the FoV, as angular topologies are notoriously difficult to expand during subsequent optimization. It also applies a directional asymmetry to the \#F. Retrieving a candidate with a smaller aperture than the target incurs a strict penalty because expanding an aperture introduces severe edge aberrations, whereas a candidate with a larger aperture is penalized lightly since stopping down the aperture is optically safe and physically stable. Overall, this strategy strictly enforces optical similarity, guaranteeing that the retrieved initial seeds are intrinsically aligned with the physical demands of the target query.

\subsection{Self-Evolving for Optical Lens Design}
\label{sec:self-evolving}

Following the retrieval of seed structures, the design agent refines the configuration through a highly integrated pipeline comprising an optical tool suite, macro orchestration, and an optics-aware self-evolving mechanism.

\paragraph{Optical Tool Suite.}~Referring to the manual optical lens design workflow, we deploy a physically grounded optical tool suite. First, we geometrically scale the entire lens to match the target effective focal length (EFFL) without altering the relative aberration profile. Second, we perform single-surface parameter adjustments under strict physical guardrails, such as automatic stop recognition and dynamic \#F synchronization. To resolve severe higher-order aberrations, we intelligently divide a single positive lens into a doublet while strictly conserving the original optical power $\phi_{\text{orig}}$:
\begin{equation}
\phi_{\text{orig}} = \frac{n_d - 1}{R_{\text{orig}}}, \quad \phi_1 = r \cdot \phi_{\text{orig}}, \quad \phi_2 = (1-r)\cdot\phi_{\text{orig}}, \quad R_i = \frac{n_d-1}{\phi_i}
\label{eq:split_lens}
\end{equation}
where $n_d$ is the material refractive index, $R_{\text{orig}}$ is the original surface radius, $r$ defines the power distribution ratio, $\phi_1$ and $\phi_2$ are the resulting split optical powers, and $R_i$ denotes the new respective radii. For continuous refinement, we apply a finite-difference gradient descent algorithm. To prevent this local optimization from sacrificing macroscopic specifications for localized image sharpness, the objective function explicitly penalizes EFFL deviation:
\begin{equation}
\widetilde{\text{RMS}}(\mathbf{R}) = \text{RMS}(\mathbf{R}) + \beta \cdot \text{RMS}_0 \cdot \max \left( 0, \frac{|\text{EFFL} - \text{EFFL}_0|}{|\text{EFFL}_0|} - \tau \right)
\label{eq:local_optimize}
\end{equation}
where $\widetilde{\text{RMS}}(\mathbf{R})$ and $\text{RMS}(\mathbf{R})$ represent the penalized objective and the current root-mean-square spot radii given the lens parameters $\mathbf{R}$. Furthermore, $\text{RMS}_0$ and $\text{EFFL}_0$ indicate the initial spot radius and target focal length, $\beta$ is the penalty weight, and $\tau$ defines the allowable tolerance threshold. Finally, we execute rigorous Damped Least Squares (DLS)~\cite{meiron1965damped} ray tracing to guarantee the physical validity of the final design.

\paragraph{Macro Orchestration.}~Instead of relying on open-ended reasoning paradigms, we orchestrate the agent using a deterministic standard operating procedure embedded within the system prompt. This strategy strictly enforces a logical design sequence, guiding the agent systematically from initial candidate retrieval to macroscopic alignment, localized modification, and final verification. To optimize computational efficiency, we introduce a two-tier evaluation mechanism. The agent first applies rapid paraxial approximations to instantly filter out configurations with extreme EFFL or \#F deviations. Only designs passing this preliminary gate proceed to the computationally expensive \texttt{Zemax} ray tracing. Furthermore, to guarantee workflow stability, a sliding window circuit breaker continuously monitors tool trajectories, actively halting the process if identical action sequences repeat to prevent infinite loops.

\paragraph{Optics-Aware Self-Evolving Mechanism}~To emulate the experience accumulation process of manual optical lens design, we employ a dual-agent architecture comprising a curriculum agent and a design agent, as shown in Fig~\ref{fig:pipeline}. In the self-evolving process, the curriculum agent generates design queries with progressively increasing difficulty, advancing from conventional specifications to challenging optical parameters. The design agent then attempts to solve these tasks, autonomously distilling and accumulating strategies across the curriculum. During each session, the design agent records detailed trajectories that explicitly link geometric actions to their corresponding physical metric changes, such as fluctuations in the RMS spot radius and EFFL. Importantly, the evolving mechanism extracts insights from both successful and failed attempts. By analyzing successful resolutions of difficult queries alongside critical failures (e.g. invalid material substitutions or unfeasible scaling limits), the agent learns both advanced design pathways and valuable physical boundary conditions. To prevent storing trivial textbook knowledge, we use a knowledge-grounded distillation prompt that extracts only novel strategies and extreme parameter adjustments. These validated skills are serialized into a persistent library and dynamically injected into future prompts, enabling \textsc{LensDesigner} to achieve continuous, zero-shot capability evolution without costly model weight updates.



\subsection{LensArena Benchmark}
\label{sec:benchmark}

In the domain of automated classical imaging lens design, recent methods~\cite{gao2025exploring,yang2024curriculum} only evaluate their performance on a handful of isolated design cases. This limited testing scope is primarily due to the immense computational overhead associated with traditional optimization pipelines. However, evaluating on restricted samples introduces randomness and bias and fails to rigorously reflect the generalization capabilities of these algorithms. Currently, the field lacks a large-scale, comprehensive testing baseline. Hence, we introduce our {LensArena} benchmark.



Our LensArena focuses specifically on classical imaging lens design. Since commonly used optical systems are highly mature and easily solved, modern engineering increasingly focuses on more challenging and extreme configurations like ultra-wide FoVs and large apertures. To rigorously test the limits of state-of-the-art automated methods against these frontiers, we structured our benchmark to reflect these modern demands. Given that full physical optical optimization is exceptionally time-consuming, our carefully curated set of 120 unique design queries constitutes a remarkably comprehensive evaluation suite. Specifically, as shown in Fig.~\ref{fig:lenslib}, LensArena deliberately includes a high proportion of challenging designs, covering a broad and diverse spectrum of specifications with the FoV ranging from $5^\circ$ to $180^\circ$ and the \#F spanning from 0.7 to 8.0. Notably, this benchmark is exceptionally challenging, as \textit{70\%} of the design queries are \textit{out-of-distribution} relative to LensLib100K. Furthermore, the benchmark queries are systematically categorized into three distinct difficulty levels. For each specific task, we define a target RMS spot radius threshold dictated by its intrinsic difficulty. A generated lens configuration is recorded as a success strictly if its final optimized radius falls below this predefined standard. To the best of our knowledge, LensArena stands as the first benchmark dedicated to the systematic evaluation of automated optical system design. For examples from the LensArena benchmark, please kindly refer to Appendix~\ref{sec:benchmark_examples}

\section{Experiments}
\label{sec:experiments}

\begin{table}[t]
\caption{
Main results evaluated on the LensArena Benchmark.
}
\label{table:lensarena-results}
\vspace{0.25em}
\centering
\small 
\renewcommand{\arraystretch}{0.85} 
\begin{tabular*}{\textwidth}{@{\extracolsep{\fill}}lcccccc@{}}
    \toprule
    & \multicolumn{3}{c}{Success Rate (\%)}~$\uparrow$ & \multicolumn{3}{c}{Avg RMS ($\mu m$)} $\downarrow$\\
    \cmidrule(lr){2-4} \cmidrule(lr){5-7}
    Method & Easy & Medium & Hard & Easy & Medium & Hard \\
    \cmidrule(lr){1-1} \cmidrule(lr){2-4} \cmidrule(lr){5-7}
    \pub{Existing Baselines}         & & & & & & \\
    \quad LensNet~\cite{cote2021deep}     & 50.0 & 39.3 & 30.0 & 3.89 & 8.99 & 7.99 \\
    \quad QGSO~\cite{gao2025exploring} & 25.0 & 22.7 & 13.3 & 24.2 & 17.7 & 11.1 \\
    \quad DeepLens~\cite{yang2024curriculum} & 4.2 & 1.5 & 0 & 29.6 & 41.1 & - \\
    \arrayrulecolor{gray}\cdashline{1-7}\arrayrulecolor{black}
    \pub{In Context Learning Baselines}       & & & & & & \\
    \quad \texttt{Gemini-3.0-flash}  &37.5 &36.3 & 26.7& 43.56& 94.7& 120.4\\
    \quad \texttt{ChatGPT-5.2}       &33.3 &27.3 &26.7 & 52.68& 90.85& 112.4\\
    \quad \texttt{Kimi-2.4}          &33.3 & 16.6& 23.3& 44.7&87.45 &108.7 \\
    \quad \texttt{Qwen-3-8b}         &8.3 &12.1 &20 &96.32 &147.7 &215.3 \\
    \quad \texttt{Qwen-3-32b}        &16.6 &21.2 &20 &67.83 & 124.23&176.43 \\
    \arrayrulecolor{gray}\cdashline{1-7}\arrayrulecolor{black}
    \textbf{\textsc{LensDesigner} (Ours)}     & & & & & & \\
    \quad \textit{w/} \texttt{Gemini-3.0-flash} & 87.5 & 83.3 & 80.0 & 3.7 & 7.5  & 12.7 \\
    \quad \textit{w/} \texttt{ChatGPT-5.2}    & 87.5 & 83.3 & 66.7 & 6.5 & 6.9 & 14.8 \\
    \quad \textit{w/} \texttt{Kimi-2.4}       & 87.5 & 74.2 & 56.7 & 9.0 & 7.8 & 11.0 \\
    \quad \textit{w/} \texttt{Qwen-3-8b}      & 70.8 & 51.5 & 40.0 & 6.5 & 9.2 & 14.8 \\
    \quad \textit{w/} \texttt{Qwen-3-32b}     & 75.0 & 75.8 & 70.0 & 4.9 & 8.2 & 7.8 \\
    \bottomrule
\end{tabular*}
\end{table}

\subsection{Experimental Setup}
\label{sec:setup}

\paragraph{Implemental Details.}~We leverage \texttt{Gemini-3.0-flash}~\cite{gemini2025}, \texttt{Kimi-2.4}~\cite{moonshot_kimi_2024}, \texttt{ChatGPT-5.2}~\cite{openai52}, and \texttt{Qwen3-32b}~\cite{yang2025qwen3} as the foundation LLMs, respectively, and
Gemini-3-flash~\cite{gemini2025} for the optics-aware retrieval reranker
and the self-evolve distiller, with lens-spec embeddings produced by
text2vec-base~\cite{text2vec} on a FAISS index. All LLM temperatures
are set to 0 to ensure reproducibility. Our optical simulator couples an
in-house paraxial ray-tracer (for offline Optics-Aware Retrieval labeling, online verification,
and gradient-based local optimize) with \texttt{Ansys Zemax OpticStudio}
\cite{zemax} accessed through ZOS-API/pythonnet behind a Flask bridge. \texttt{Zemax}
is the sole source of truth for final RMS / EFFL / \#F. See
Appendix~\ref{supp:implementation} for more details

\paragraph{Evaluation Metrics.}~Given the absence of prior similar works in this domain, we formally define three primary metrics to evaluate the generated lens designs. The \textit{Success Rate} measures the percentage of tasks where the final RMS spot radius successfully reaches a predefined target threshold, which varies dynamically based on the specific problem difficulty. 
The \textit{Avg RMS} denotes the average RMS spot radius, sampled at 0, 0.7, and 1.0 times the maximum FoV. As the standard metric for optical evaluation, a smaller RMS indicates tighter ray convergence and superior image quality. Note that the \textit{Avg RMS} is averaged only over the subset of successful designs, excluding failed queries.


\paragraph{Baselines}~To comprehensively evaluate our proposed approach, we compare it against three publicly available automated lens design methods and specifically constructed language model baselines. {LensNet}~\cite{cote2021deep} serves as a data-driven baseline, utilizing a database retrieval mechanism combined with simple geometric scaling. {QGSO}~\cite{gao2025exploring} is an optimization-based approach powered by genetic algorithms. It stands as the current SOTA in automated optical design and can generate most conventional lens structures, though it suffers from exceedingly slow optimization speeds. {DeepLens}~\cite{yang2024curriculum} represents a deep-learning-based framework that employs curriculum learning strategies to formulate lens configurations. Given the complete absence of existing LLM-based frameworks for this specific domain, we introduce LLM-based in-context-learning (ICL) baselines. This method relies on standard in-context learning and is driven by the exact same foundation language model as our framework to ensure a rigorously fair comparison.

\begin{figure}
    \centering
    \includegraphics[width=0.97\linewidth]{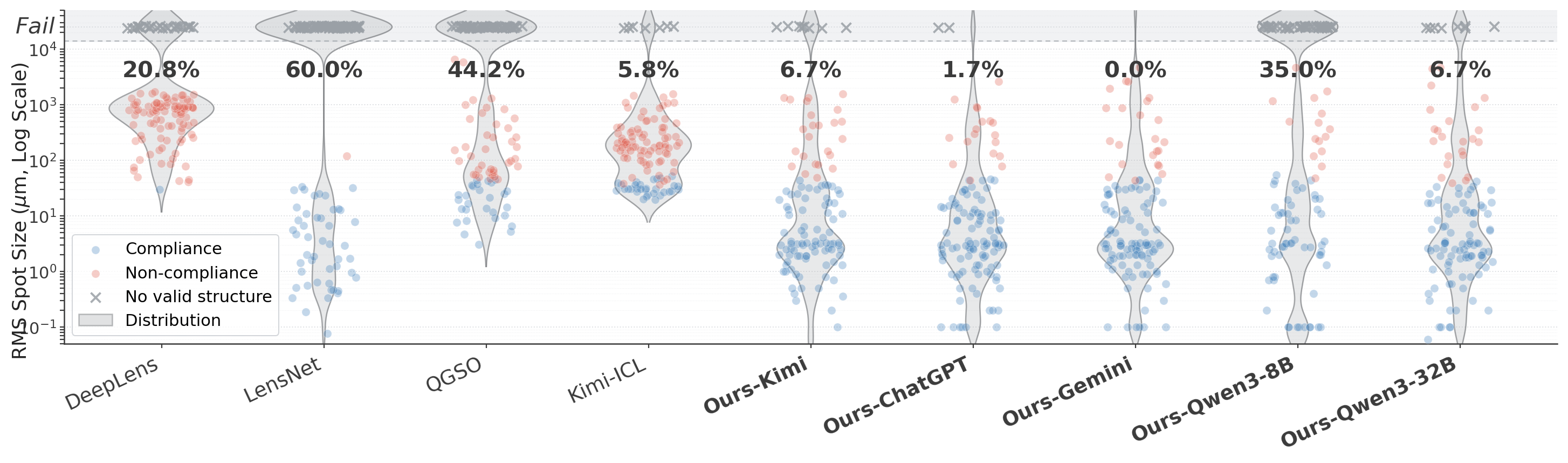}
    \vspace{-10pt}
    \caption{Results on the LensArena benchmark. The evaluation demonstrates that our proposed method achieves superior structural rationality, a higher success rate, and a significantly lower RMS spot radius compared to the baselines.}
    \label{fig:benchmark}
\end{figure}

\subsection{Results}
\label{sec:results}

\paragraph{Comparison on the LensArena Benchmark.}
Tab.~\ref{table:lensarena-results} presents the quantitative evaluation of \textsc{LensDesigner} against existing computational baselines and ICL approaches across three difficulty levels. Compared to existing deep learning and heuristic optimization baselines, our proposed framework demonstrates a substantial leap in both capability and design robustness. Specifically, powered by \texttt{Gemini-3.0-flash}, \textsc{LensDesigner} achieves a remarkable success rate of 87.5\% on the Easy split and maintains a highly robust 80.0\% success rate even on the Hard split. In stark contrast, prior methods struggle significantly as the problem complexity increases; their success rates on Hard tasks degrade to 30.0\% at best (LensNet) or fail completely (0\% for DeepLens). Furthermore, while conventional optimization methods typically require several hours to explore the complex, non-convex parameter space, our agentic framework converges to high-quality optical solutions (e.g., yielding an Avg RMS of just 12.7 $\mu m$ on Hard queries) in merely tens of seconds.

As shown in Fig.~\ref{fig:benchmark}, since LensNet is a retrieval-based method, it achieves a relatively high success rate for structures covered by its database. However, $60\%$ of its generated structures fail during optical ray tracing because they fall outside the scope of its library. Conversely, although DeepLens produces valid structures in approximately $80\%$ of cases, its final performance compliance rate is notably low. In contrast, our proposed method consistently yields a vast majority of structurally valid designs while achieving a superior compliance rate for the required RMS spot radius.

Compared to LLM-based baselines, relying solely on open-ended, prompt-based ICL proves highly ineffective for stringent engineering tasks. As shown in Tab.~\ref{table:lensarena-results}, standard ICL approaches suffer from remarkably low success rates and massive RMS errors across all difficulty tiers due to frequent physical violations and poor convergence. In contrast, our fully equipped \textsc{LensDesigner} pipeline consistently and significantly outperforms its corresponding ICL counterparts across every metric. This substantial performance margin confirms that integrating domain-specific tools, optics-aware retrieval, and systematic macro orchestration is essential to unlock the true potential of foundation models in physical optical design.

\begin{figure}
    \centering
    \includegraphics[width=0.98\linewidth]{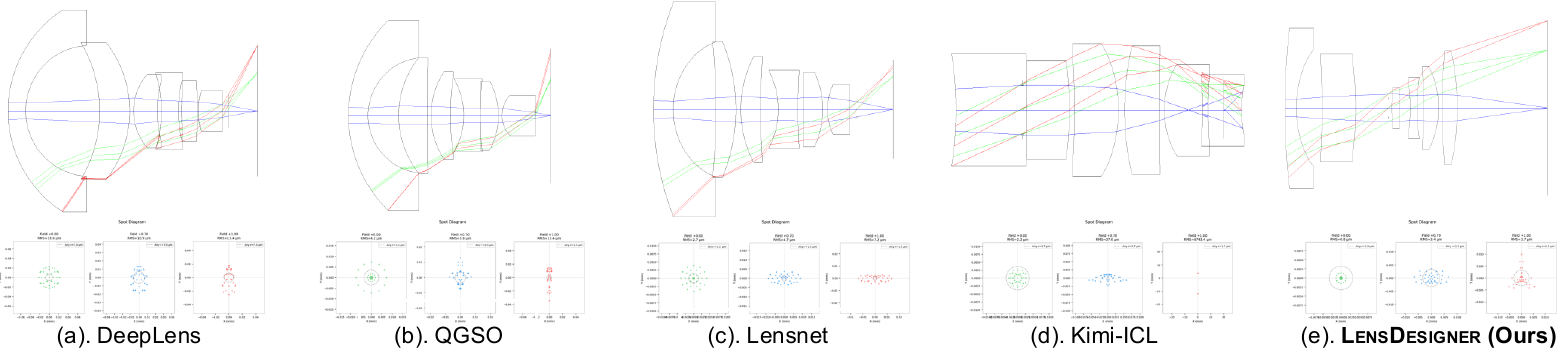}
    \vspace{-5pt}
    \caption{Visualized results of lens layouts and spot diagrams from the LensArena benchmark. \textsc{LensDesigner} consistently yields optimal structural configurations and achieves the minimal RMS spot size, outperforming other methods in both design rationality and imaging quality.}
    \label{fig:qualitative}
    \vspace{-12pt}
\end{figure}

\paragraph{Example Analysis.}
As the lens layouts and the corresponding RMS spot diagrams shown in Fig.~\ref{fig:qualitative}, \textsc{LensDesigner} generates highly practical and physically valid optical structures. The layout visualization demonstrates a rational distribution of optical power across the generated glass elements, effectively managing internal ray angles to minimize spherical aberrations and chromatic dispersion. Furthermore, the accompanying spot diagrams confirm exceptional ray convergence at the focal plane. The tightly concentrated focal spots indicate that the generated designs strictly satisfy the demanding imaging specifications. This superior optical performance directly validates the effectiveness of our \textsc{LensDesigner}.

\begin{wrapfigure}{l}{0.5\textwidth}
    \centering
    \includegraphics[width=\linewidth]{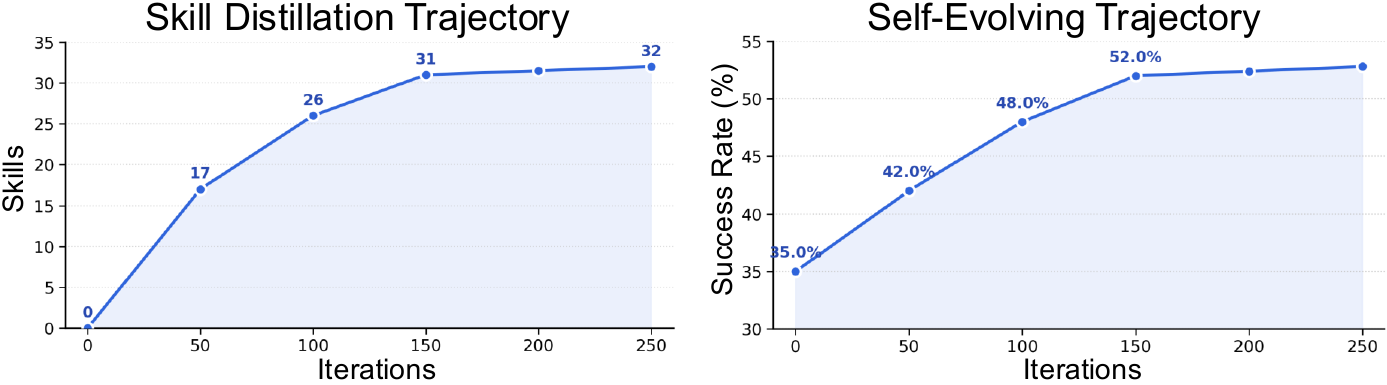}
    \caption{Trajectory of self-evolving.}
    \label{fig:trajectory}
    \vspace{-8pt}
\end{wrapfigure}

\paragraph{Self-Envolving.}
To empirically validate the continuous learning capability of our framework, we track and visualize the distilled skills and success rate across progressive self-evolving iterations in Fig~\ref{fig:trajectory}. As illustrated, as the skills are accumulated over iterations, the overall success rate exhibits a steady and significant upward trajectory as the number of iterations increases. This definitive learning curve directly proves the efficacy of our optics-aware evolving mechanism. It demonstrates that \textsc{LensDesigner} is not merely executing static rule-based procedures, but is actively and effectively exploring the highly complex optical design space. By continuously distilling advanced strategies from both novel successes and boundary-testing failures, the agent progressively elevates its intrinsic optical engineering proficiency. 


\subsection{Ablation Study}
\label{sec:ablation}

To validate the effectiveness of each core module within \textsc{LensDesigner}, we conduct a comprehensive ablation study using \texttt{Gemini-2.5}~\cite{google_gemini_2_5}, as summarized in Table~\ref{table:ablation}. 

\textbf{Optics-Aware Retrieval:} Removing \texttt{LensLib100K} entirely forces the agent to design from scratch, leading to the worst overall performance. Furthermore, replacing our optics-aware retrieval with a Vanilla RAG yields sub-optimal initial seeds, resulting in a severe degradation of the RMS spot radius, particularly on hard queries (189.6 $\mu m$ vs. 48.8 $\mu m$). This demonstrates the necessity of matching candidates based on deterministic physical metrics rather than mere semantic proximity.

\textbf{Tool Suite \& Macro Orchestration:} Ablating the Optical Tool Suite or Macro Orchestration significantly reduces the success rate and increases aberration errors. This indicates that relying solely on the open-ended reasoning of large language models is insufficient for optical design. Strict physical guardrails and systematic, deterministic workflows are essential for continuous convergence.

\textbf{Self-Evolving Mechanism:} We further ablate the components of the self-evolving module. Removing either the reflection mechanism (which analyzes errors) or the skill library (which stores persistent cross-task heuristics) leads to notable performance drops on medium and hard tasks. This confirms that dynamically distilling and transferring historical experience is critical for mastering complex optical specifications. Overall, the full \textsc{LensDesigner} pipeline achieves the highest success rates and the lowest Avg RMS across all difficulty tiers.

\begin{figure}
    \centering
    \includegraphics[width=0.95\linewidth]{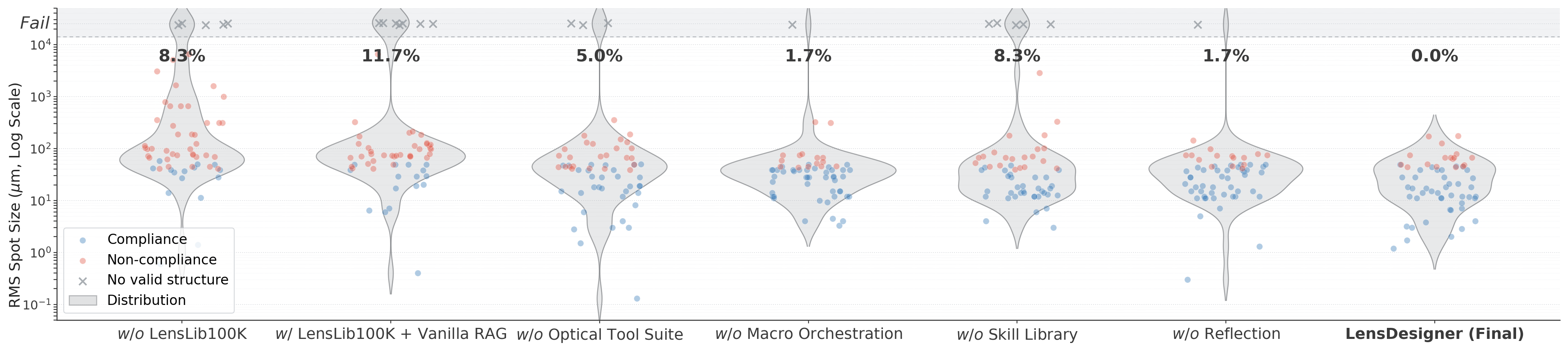}
    \vspace{-8pt}
    \caption{Ablation study on the core components of \textsc{LensDesigner}. The violin plots illustrate the distribution of RMS spot radii across different model variants on the LensArena benchmark.}
    \label{fig:ablation}
    \vspace{-8pt}
\end{figure}

\begin{table}[t]
\caption{
Ablation study on the core components of \textsc{LensDesigner} using the LensArena Benchmark. All variants are evaluated on the identical foundation model for fair comparison.
}
\label{table:ablation}
\vspace{0.25em}
\centering
\small 
\renewcommand{\arraystretch}{0.95} 
\begin{tabular*}{\textwidth}{@{\extracolsep{\fill}}lcccccc@{}}
    \toprule
    & \multicolumn{3}{c}{Success Rate (\%)} $\uparrow$ & \multicolumn{3}{c}{Avg RMS ($\mu m$)} $\downarrow$ \\
    \cmidrule(lr){2-4} \cmidrule(lr){5-7}
    Method & Easy & Medium & Hard & Easy & Medium & Hard \\
    \cmidrule(lr){1-1} \cmidrule(lr){2-4} \cmidrule(lr){5-7}
    \textit{w/o} LensLib100K                & 25.0 & 39.3 & 26.7 & 89.6 & 194.4 & 293.4 \\
    \textit{w/} LensLib100K + Vanilla RAG        & 58.3 & 51.5 & 40.0 & 35.1 & 128.1 & 189.6 \\
    \textit{w/o} Optical Tool Suite                              & 66.7 & 63.6 & 26.7 & 37.1 & 54.7  & 68.9 \\
    \textit{w/o} Macro Orchestration    & 80.0 & 69.7 & 33.3 & 21.8 & 34.2  & 87.9 \\
    \arrayrulecolor{gray}\cdashline{1-7}\arrayrulecolor{black}
    \pub{Self-Envolving}       & & & & & & \\
    \textit{w/o} Reflection      & 66.7 & 60.6 & 46.7 & 26.1 & 31.2  & 81.4 \\
    \textit{w/o} Skill Library & 66.7 & 72.7 & 46.7 & 23.2 & 31.2  & 81.4 \\
    \arrayrulecolor{gray}\cdashline{1-7}\arrayrulecolor{black}
    \textbf{\textsc{LensDesigner} (Final)} & 91.7 & 70.0 & 53.2 & 19.2 & 33.4 & 48.8 \\
    \bottomrule
\end{tabular*}
\vspace{-8pt}
\end{table}

\section{Conclusion}
\label{sec:conclusion}

In this paper, we introduced \textsc{LensDesigner}, the first large language model-driven agent framework for automated classical imaging lens design. To comprehensively evaluate performance in this domain, we also proposed LensArena, the first large-scale and systematic benchmark tailored for optical system design. Extensive experiments demonstrate that our approach significantly outperforms existing optimization-based and deep-learning-based methods in both success rate and design efficiency, particularly on complex tasks with extreme optical parameters.
\vspace{-5pt}

\paragraph{Limitations and Future Work.}
Currently limited to spherical lenses, \textsc{LensDesigner} faces challenges in scaling to aspheric or freeform surfaces and suffers from latency due to external ray-tracing software. Future efforts will extend its capabilities to non-spherical and complex folded systems to further automate optical engineering.

\newpage
\bibliography{reference}
\bibliographystyle{plain}

\clearpage

\newpage

\appendix

\section*{Appendix}

\section{Detailed Optical Design Preliminaries}
\label{supp:preliminaries}

\begin{figure}[h]
    \centering
    \includegraphics[width=0.8\linewidth]{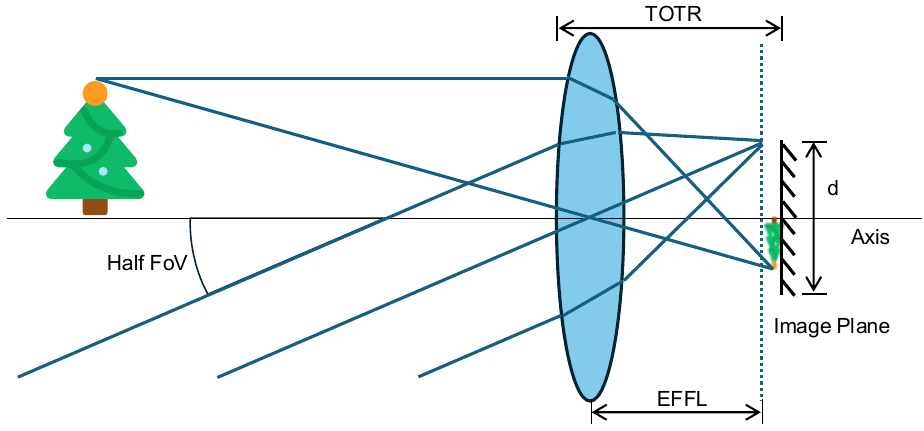}
    \caption{A simple example of a lens system.}
    \label{fig:pre_supp}
\end{figure}

This section provides a comprehensive exposition of the fundamental optical principles, the physical imaging process, and the quantitative metrics essential for evaluating lens configurations within our proposed framework. A simple example of lens system is shown in Fig.~\ref{fig:pre_supp}. 

\subsection{The Physical Imaging Process and Essential Components}
The formation of an optical image is a deterministic physical process governed by the laws of refraction. In a standard computational environment, this process is modeled via sequential ray tracing. Light rays propagate from a source target, traverse multiple refractive media, and ultimately converge to form a visible representation. A complete spherical lens system necessitates the following fundamental structural elements:
\begin{itemize}
    \item \textbf{Object Plane:} The physical or virtual surface where the target scene originates. Light rays radiate from discrete coordinate points on this plane toward the optical system.
    \item \textbf{Optical Elements:} A sequence of spherical lenses manufactured from specific glass materials. Each individual element is mathematically defined by its front and rear surface curvatures, central geometric thickness, and distinct material properties, specifically the refractive index and Abbe number. These elements sequentially refract incoming light bundles to precisely manipulate their trajectory.
    \item \textbf{Stops and Pupils:} The aperture stop is a physical boundary that restricts the diameter of the light bundle traversing the system. It governs the overall illumination brightness and depth of field. The effective optical images of this aperture stop, as formed by the lenses preceding and following it, are formally known as the entrance pupil and exit pupil, respectively.
    \item \textbf{Image Plane:} The terminal spatial surface where the refracted light rays intersect to form the final focused picture. In modern digital optics, this physical plane strictly coincides with the location of the electronic image sensor.
\end{itemize}

\subsection{Detailed Explanation of Optical Parameters}
To rigorously constrain the design parameter space and quantitatively evaluate optimization outcomes, we rely on a specific set of optical metrics. A detailed theoretical explanation of each metric is provided below:

\begin{itemize}
    \item \textbf{Field of View (FoV):} This metric quantifies the maximum angular extent of the external environment that the optical system can project onto the sensor. It directly dictates the observational breadth of the lens. A larger FoV corresponds to wide angle photography, whereas a narrow FoV is characteristic of telephoto systems specifically designed for distant target observation.
    \item \textbf{F number (\#F):} Defined as the dimensionless ratio of the Effective Focal Length to the physical diameter of the entrance pupil. It serves as a universal indicator of the optical aperture size and the overall light gathering capability. A smaller F number indicates a comparatively larger aperture, allowing substantially more light to reach the sensor, which is highly beneficial for low illumination environments.
    \item \textbf{Optical Format ($d$) and Semi Image Height ($y$):} The optical format refers to the physical diagonal dimension of the active image sensor. To prevent mechanical vignetting and ensure full sensor coverage, the optical system must project an image circle that completely encompasses this sensor. The semi-image height $y$ is exactly half of the optical format ($y = d / 2$), representing the maximum radial coordinate on the image plane that requires sharp optical resolution.
    \item \textbf{Effective Focal Length (EFFL):} The physical distance measured from the rear principal plane of the optical system to the focal point on the image plane. It is the primary determinant of optical magnification. For a fixed sensor size, the EFFL mathematically restricts the Field of View, establishing a fundamental trade-off in optical geometry.
    \item \textbf{Total Track (TOTR):} This absolute metric measures the physical length of the entire optical assembly, calculated along the central optical axis from the front vertex of the first lens element to the final image plane. It is a critical geometric constraint for compact device packaging, such as integrating camera modules into thin mobile devices.
    \item \textbf{Root Mean Square (RMS) Spot Radius:} The most critical quantitative metric for evaluating optical aberrations. When light rays from a single infinitesimal point on the object plane pass through a non ideal lens system, they do not converge perfectly to a single identical point on the image plane. Instead, they scatter mathematically to form a microscopic blur circle known as the focal spot. The RMS computes the standard deviation of the spatial distribution of these intersecting rays. A smaller RMS spot radius directly correlates to higher imaging sharpness, minimal structural aberrations, and superior spatial resolution.
\end{itemize}

\section{Detailed Related Work}
\label{supp:related_work}

\paragraph{Automatic Lens Design.}~
As previously noted, optical lens design is an empirical task highly dependent on human expertise. While modern software tools like Zemax and CODE V provide powerful simulation environments, they function primarily as calculators for intensive numerical optimization rather than autonomous decision makers.
To alleviate the heavy reliance on manual labor and expert intuition, researchers have proposed various automatic lens design algorithms. These approaches primarily include methods driven by continuous optimization such as QGSO~\cite{guo2019new, zhang2020automated, gao2025exploring, liu2025global, gao2025neuro, yang2017automated, zhang2021towards}, neural network models including DeepLens~\cite{cote2019extrapolating, cote2021deep, yang2024curriculum}, and database retrieval techniques like LensNet~\cite{cote2021deep}.
However, these existing solutions still present significant practical limitations, which often demand meticulous manual intervention and parameter tuning, frequently requiring several days to converge on a viable solution. Furthermore, pure retrieval techniques are inherently constrained and cannot handle novel structural requirements outside their existing libraries. In contrast, our proposed framework aims to rapidly generate compliant lens configurations through simple text interaction, entirely circumventing the need for complex parameter adjustments, specialized domain expertise, and prolonged optimization periods.

\paragraph{LLMs and LLM-based Agents.}
Massive language models are increasingly recognized as the primary foundation for achieving artificial general intelligence~\cite{wei2022emergentabilitieslargelanguage,bubeck2023sparksartificialgeneralintelligence,bommasani2022opportunitiesrisksfoundationmodels}. Moving past traditional capacity expansion principles, the latest generative architectures, such as LLaMA~\cite{touvron2023llama}, DeepSeek~\cite{liu2024deepseek}, Qwen3~\cite{yang2025qwen3}, Claude Sonnet~\cite{anthropic2024claude}, and GPT 5~\cite{singh2025openai}, clearly demonstrate their capability to serve as the central reasoning mechanisms for independent automated entities~\cite{xi2023risepotentiallargelanguage,Wang_2024}.
By encapsulating cognitive engines within interactive frameworks, intelligent agents autonomously perceive contexts, formulate plans, and execute complex actions to significantly elevate productivity. For instance, OpenClaw~\cite{openclaw2026} leverages tool utilization and environmental feedback for precise robotic manipulation. Similarly, these cognitive agents are revolutionizing other specialized disciplines, ranging from ChemCrow~\cite{bran2023chemcrow} designing chemical synthesis pathways to software agents~\cite{yang2024swe,zhang2024autocoderover,qian2024chatdev} resolving intricate programming anomalies. Despite these interdisciplinary triumphs, deploying such autonomous entities within the highly empirical and numerically demanding domain of optical lens design remains largely unexplored~\cite{geng2026optiagentphysicsdrivenagenticframework}.

\section{Detailed Optics-Aware Retrieval Mechanism}
\label{supp:retrieval}
This section elaborates on our Optics-Aware Retrieval module. Rather than merely returning similar documents, our retrieval framework operates as a physical performance driven selector and an Out of Distribution router. We bypass conventional semantic matching by incorporating three domain specific innovations.

\paragraph{More Details about LensLib100K Construction}
We use the enhanced QGSO~\cite{gao2025exploring} automatic optical design algorithm. QGSO first define a $Population$ composed of all lens structures to be optimized, where each structure is coined an $Individual$. 
Without any empirical preference, the $Individual$ is the normalized lens parameters vector $P$ from random initialization, and $m$ $Individuals$ are initialized to constitute the $Population$ for enriching diversity.
Then, a hybrid global and local optimization strategy is proposed to find multiple promising high-quality structures $\hat{\mathrm{P}}$ for $\mathrm{P}$, coined the $Parent$, which minimizes the Merit Function.
In addition, $Parent$ is mutated for more possible structures, the mutated $Parent$ is then mixed with the origin $Parent$ to serve as the start points of the next generation.
Finally, the hybrid optimization, selection, and mutation processes are re-conducted for each generation, where the $Parent$ is outputted as the diverse lens structures when the generation meets the set number.

\paragraph{Simulator in the Loop Labeling}
Standard retrieval augmented generation relies on semantic text embeddings. Conversely, we perform physical characterization on every lens before adding it to the database. Utilizing an offline paraxial ray tracer, we evaluate each structure across multiple field coordinates to precompute essential physical metrics, including the Effective Focal Length, total track length, paraxial root mean square spot radius, and actual aperture radius. Consequently, candidate selection is solidly grounded in deterministic physical states rather than heuristic language estimations.

\paragraph{Asymmetric Optical Prior Distance}
The absolute core of our retrieval engine is an asymmetrical distance metric designed strictly around optical engineering realities. The distance $d$ between a candidate lens and the target specification is formulated as:
$$d = w_{\text{FoV}} \cdot \left| \frac{\text{FoV}_{\text{lens}} - \text{FoV}_{\text{tgt}}}{\text{FoV}_{\text{tgt}}} \right| + w_{\text{\#F}}^{\pm} \cdot \left| \frac{\text{\#F}_{\text{lens}} - \text{\#F}_{\text{tgt}}}{\text{\#F}_{\text{tgt}}} \right| + w_{\text{RMS}} \cdot \text{RMS}_{\text{lens}}$$
We assign distinct numerical weights based on absolute physical constraints. The FoV carries a massive symmetric weight of 10.0 because an optical topology is fundamentally restricted by its maximum angular acceptance, making post optimization expansion virtually impossible. Conversely, the \#F weight $w_{\text{\#F}}^{\pm}$ is heavily asymmetric. If the candidate \#F is larger than the target (indicating a smaller aperture), we apply a strict penalty of 1.0. Expanding an aperture physically exposes the system to severe marginal aberrations, carrying immense optimization risk. However, if the candidate \#F is smaller (a larger aperture), we apply a highly lenient penalty of 0.5, because shrinking an aperture reliably eliminates aberrations with near zero risk. The paraxial root mean square radius serves merely as a tiebreaker with a minimal weight of 0.01 to prevent early overfitting to paraxial approximations.

\paragraph{Two Stage Retrieval and Active Routing}
Our framework utilizes a widely adopted indexing database for initial screening, but pure numerical distance does not consistently guarantee the easiest trajectory. Therefore, we implement a two stage pipeline. First, the distance formula extracts the top twenty candidates. Next, a lightweight language model evaluates these candidates using structural heuristics to select the final top five starting structures. Furthermore, our retrieval module acts as an active router. If the target specifications represent an Out of Distribution query (specifically defined as a FoV deviation exceeding 10° or an \#F deviation exceeding 0.5), the system explicitly attaches a Standard Operating Procedure to the prompt. This proactive routing instructs the downstream agent on the exact sequential recovery path required, significantly enhancing robustness in sparsely populated optimization spaces.

\section{Detailed Methodology}
\label{supp:methodology}

This section provides a comprehensive technical breakdown of the autonomous optical design framework, detailing the exact physical equations, algorithmic guardrails, and prompt engineering strategies utilized to ensure physical validity and deterministic convergence.

\subsection{Optical Tool Suite Implementation}
The core of our framework relies on an optical tool suite. Unlike generic numerical editors, each tool strictly embeds optical engineering principles and boundary conditions.

\paragraph{Deterministic Focal Length Alignment (\texttt{align\_effl}).}
When aligning the effective focal length (EFFL) of a retrieved candidate to the user specification, we avoid using optimization search. Instead, we apply a deterministic geometric scaling. All surface radii, thicknesses, and semi-diameters are synchronously multiplied by a scale factor defined as $\text{scale} = \text{EFFL}_{\text{target}} / \text{EFFL}_{\text{current}}$. This transformation precisely scales the EFFL linearly while preserving the inherent relative aberration profiles. To prevent unphysical expansions, we enforce a strict 300\% scaling upper limit. Crucially, the updated EFFL is immediately written back to the internal model state to prevent downstream evaluators from targeting outdated specifications.

\paragraph{Guarded Single-Surface Modification (\texttt{modify\_lens}).}
To facilitate localized adjustments, this tool alters single parameters while being constrained by five distinct physical guardrails to prevent fatal execution errors, as summarized in Table~\ref{tab:guardrails}.

\begin{table}[h]
\centering
\caption{Physical guardrails implemented within the \texttt{modify\_lens} tool.}
\label{tab:guardrails}
\begin{tabular}{ll}
\toprule
\textbf{Guardrail Type} & \textbf{Trigger Condition and Action} \\
\midrule
Stop Surface Recognition & Rejects semi-diameter edits on non-stop surfaces. \\
Aperture Scaling Limits  & Rejects semi-diameter scaling outside the $[0.5, 3.0]$ ratio. \\
Dynamic F-number Sync    & Synchronizes system F-number automatically after aperture edits. \\
Material Whitelist       & Rejects non-catalog glasses (restricted to verified H- and D- types). \\
Loop Prevention          & Rejects identical edits repeated $\geq 3$ times, prompting a strategy shift. \\
\bottomrule
\end{tabular}
\end{table}

\paragraph{Optical Power-Conserving Split (\texttt{split\_lens}).}
To mitigate severe higher-order spherical aberrations induced by excessive local curvature, we introduce a tool that splits a single positive lens into two separate elements. This operation rigorously conserves the original optical power $\phi_{\text{orig}}$ derived from the thin-lens approximation:
\begin{equation}
\phi_{\text{orig}} = \frac{n_d - 1}{R_{\text{orig}}}, \quad \phi_1 = r \cdot \phi_{\text{orig}}, \quad \phi_2 = (1-r)\cdot\phi_{\text{orig}}, \quad R_i = \frac{n_d-1}{\phi_i}
\end{equation}
The power distribution ratio $r$ is bounded within $[0.2, 0.8]$ (defaulting to 0.5), ensuring a balanced division of the refractive workload. 

\paragraph{Penalized Local Optimization (\texttt{local\_optimize}).}
We implement a custom finite-difference gradient descent algorithm to provide rapid, localized fine-tuning. A critical innovation here is the integration of an EFFL soft penalty directly into the objective function. This ensures that the optimizer does not collapse the macroscopic focal length to artificially minimize the root-mean-square (RMS) spot radius:
\begin{equation}
\widetilde{\text{RMS}}(\mathbf{R}) = \text{RMS}(\mathbf{R}) + \beta \cdot \text{RMS}_0 \cdot \max \left( 0, \frac{|\text{EFFL} - \text{EFFL}_0|}{|\text{EFFL}_0|} - \tau \right)
\end{equation}
We empirically set the penalty weight $\beta = 0.5$ and the tolerance threshold $\tau = 0.03$. The gradient is estimated using central differences with an adaptive epsilon. To ensure geometric stability, if the updated curvature flips its sign (transitioning a convex surface to concave), the step is rejected, and the radius is clamped to 95\% of its previous value. Optimization terminates early if the RMS reaches a noise floor of $3 \, \mu m$. The \texttt{random\_restart} tool complements this by applying controlled random perturbations uniformly sampled up to 15\% to escape local minima.

\paragraph{Ground-Truth Verification (\texttt{zemax\_optimize}).}
The ultimate physical evaluation is conducted using the Damped Least Squares (DLS) algorithm within Zemax OpticStudio. To prevent the DLS algorithm from diverging or generating unmanufacturable shapes, we establish several critical pre-processing rules and a customized Merit Function. Most notably, we mandate that the central thickness of glass elements remains non-variable during DLS execution. Without this constraint, the optimizer frequently compresses the glass thickness to near zero, resulting in unmanufacturable hourglass-shaped lenses. Additionally, as detailed in Table~\ref{tab:merit_function}, we assign asymmetric weightings across different field of views (FOV) and utilize unilateral penalties for the total track length (TOTR) to strictly penalize oversized systems without punishing compact ones.

\begin{table}[h]
\centering
\caption{Custom Merit Function configuration for Zemax DLS optimization.}
\label{tab:merit_function}
\resizebox{\textwidth}{!}{
\begin{tabular}{llcl}
\toprule
\textbf{Operand} & \textbf{Target} & \textbf{Weight} & \textbf{Physical Justification} \\
\midrule
RSCE (Per Field) & 0 & 4 / 5 / 6 & Escalating weights for challenging marginal fields. \\
EFFL / WFNO      & User Spec & 1.0 / 2.0 & Weak anchoring to prevent macroscopic drift. \\
OPLT (Max TOTR)  & Max Limit & 2.0 & Unilateral penalty applied only when exceeding limits. \\
MNCG             & $2.0 \, mm$ & 50.0 & Hard constraint for minimum manufacturable glass thickness. \\
\bottomrule
\end{tabular}%
}
\end{table}

A crucial engineering component of this tool is the explicit state synchronization mechanism. Upon completion of the DLS optimization, the finalized surface geometries are immediately written back to the agent's internal memory state. This prevents the severe desynchronization bug where the agent evaluates subsequent actions using an outdated mathematical model while the external software holds the optimized state.

\subsection{Macro Orchestration and Workflow}
To navigate the expansive optical design space effectively, we replace generic open-ended reasoning paradigms with a deterministic Standard Operating Procedure (SOP) embedded in the system prompt. The agent must sequentially execute retrieval, macroscopic alignment, microscopic optimization, and final verification. 

To optimize token usage and computational overhead, we introduce a two-tier evaluation mechanism. The \texttt{check\_spec} tool acts as a rapid paraxial filter. It computes initial metrics and strictly fails any configuration exhibiting EFFL deviations greater than 2\%, or parameter combinations deemed mathematically infeasible (such as an FOV deviation exceeding 50\%). It uniquely employs asymmetric validation for the F-number, instantly passing designs with smaller-than-target F-numbers since restricting the aperture is a trivial subsequent operation. 

To mitigate infinite loops, a sliding window circuit breaker tracks the most recent ten tool actions. If any identical tool and normalized input combination repeats three times within this window, the system triggers a forced strategy shift. Furthermore, to accommodate context length limits, the prompt solely indexes available skills by name, dynamically loading the full verbose instructions only when the \texttt{get\_skill\_detail} function is actively invoked.

\subsection{Knowledge-Grounded Self-Evolving Mechanics}
Our framework possesses a zero-weight-updating self-evolving mechanism that autonomously distills expert strategies from task trajectories.

\paragraph{Causal Trajectory Recording.}
Unlike conventional text-based trajectory logging, our memory module strictly tracks physical causality. Each record explicitly juxtaposes the geometric action against the resulting state changes in RMS and EFFL. 

\paragraph{Heuristic Distillation Filters.}
To ensure only high-quality knowledge enters the skill library, we apply six specific heuristic filters before distillation. Notably, the system aggressively prioritizes trajectories that succeed on out-of-distribution (OOD) parameters where the database provides poor initial coverage. Furthermore, it explicitly distills negative experiences. For example, failing after a specific glass substitution teaches the agent valuable material boundary conditions, converting a failed session into a reusable constraint rule.

\paragraph{Optics-Aware Distillation Prompt.}
To prevent the skill library from being flooded with fundamental optical textbook principles, we employ a Optics-Aware distillation strategy. The distilling language model is explicitly provided with a baseline of common optical heuristics. It is strictly instructed to reject standard practices and extract only insights that transcend this baseline, such as specific non-intuitive operation sequences or extreme parametric combinations that yielded unexpected success.

\paragraph{Novelty Assessment and Persistence.}
Extracted skills undergo rigorous deduplication. A skill is rejected if its Bigram-Jaccard similarity score compared to existing skills exceeds 0.80. However, an identical sequence of tools is still classified as a novel skill if it employs extreme numerical parameters (such as an EFFL scaling factor exceeding 2.0 or a single-step RMS improvement exceeding 80\%). These validated, highly specific skills are serialized into a persistent JSON library and injected directly into the active prompt context for all subsequent tasks, establishing a continuous and interpretable evolution cycle.

\subsection{General Implementation Details}
\label{supp:computation_details}
The complete framework is instantiated utilizing the LangChain architecture, encompassing standard modular components for tool execution and retrieval. The reasoning trajectory is managed using a sliding window scratchpad that retains only the ten most recent steps to prevent context overflow. Communication with the Zemax OpticStudio environment is facilitated through a robust, cross-process Flask bridge, allowing the Python-based agent to seamlessly orchestrate the proprietary physical engine.

Our hardware setup includes an NVIDIA 4090 GPU (24GB VRAM) paired with an Intel Core i9-14900K CPU and 64GB of RAM. While we access most LLMs via their official APIs, the Qwen models are deployed locally and run exclusively on our GPU.

\section{More Qualitative Results}
\label{supp:qualitative}
\begin{figure}
    \centering
    \includegraphics[width=0.98\linewidth]{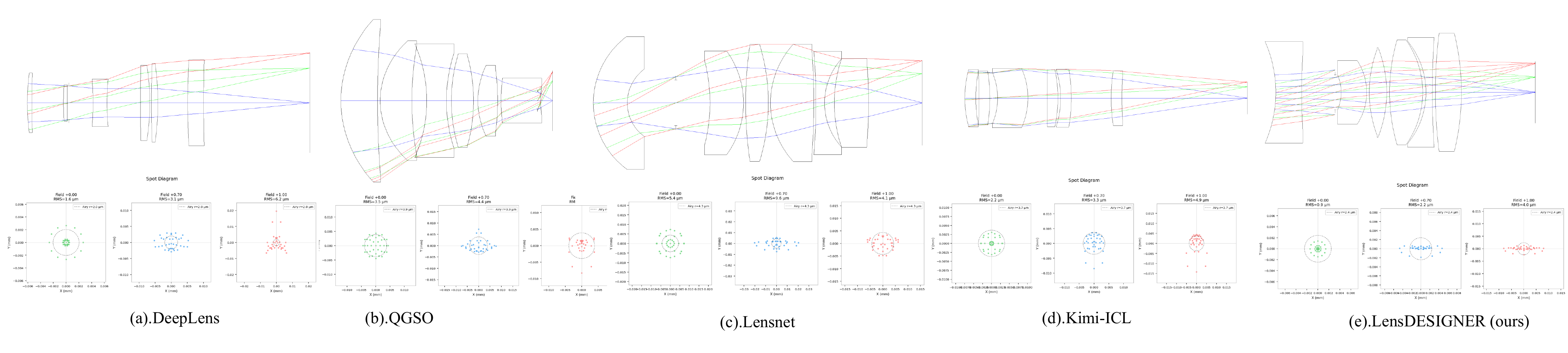}
    \caption{Additional visualized results of lens layouts and spot diagrams from the \texttt{LensArena} benchmark. \textsc{LensDesigner} demonstrates robust performance across a diverse range of design specifications, consistently producing optimized optical structures with superior imaging convergence compared to baseline methods.}
    \label{fig:qualitative_supp}
\end{figure}

Here we show more qualitative results from our LensArena benchmark in Fig.~\ref{fig:qualitative_supp}.

\section{More Experimental Setup Details}

\subsection{More Implementation Details}
\label{supp:implementation}

We use a locally served \emph{Qwen3-32b}~\cite{yang2025qwen3} (deployed via vLLM with an
OpenAI-compatible endpoint and the \texttt{enable\_thinking=False} chat-template flag)
as the primary tool-calling agent, and \emph{Gemini-3-flash-preview}~\cite{gemini2025}
for both the optics-aware retrieval reranker (Sec.~\ref{sec:lenslib}) and the
self-evolve trajectory distiller (Sec.~\ref{sec:self-evolving}). Lens-spec
embeddings are produced by the open-source \emph{text2vec-base}
model~\cite{text2vec} on a locally
persisted FAISS index. We set all LLM temperatures to 0 to ensure reproducibility
of tool selection, retrieval reranking, and skill distillation, with
\texttt{max\_tokens} of 512 for the agent loop, 1{,}000 for the reranker, and
8{,}000 for the distiller. Our optical simulator is a two-tier setup that
couples an in-house Python paraxial ray-tracer—used for offline RAG labeling,
online \texttt{check\_spec} verification, and the \texttt{local\_optimize}
finite-difference gradient descent—with \emph{Ansys Zemax OpticStudio}~\cite{zemax}
accessed through the ZOS-API .NET bindings (\texttt{pythonnet}) and wrapped
behind a Flask HTTP bridge, with Zemax serving as the sole source of truth for
final RMS / EFFL / F-number / merit-function values; all glass materials are
restricted to the CDGM library, and Zemax local optimization runs the
Damped-Least-Squares solver under our customized merit function. The ReAct controller is built on LangChain's
\texttt{create\_react\_agent} with a 10-step sliding-window scratchpad and a
maximum of 25 reasoning iterations per query.

\section{Data Examples from LensArena}
\label{sec:benchmark_examples}

Here we show some examples from our LensArena benchmark.

\input{supp/benchmark_examples}

\section{Limitations and Broader Impact}
\label{supp:limitation}

\subsection{Detailed Limitations}
While \textsc{LensDesigner} significantly advances autonomous optical engineering, several technical boundaries remain:
\begin{itemize}
    \item \textbf{Geometrical Constraints:} Our current framework is primarily optimized for co-axial spherical systems. It does not yet natively support off-axis systems, tilted components, or non-imaging optics (e.g., light guides), which require higher-dimensional spatial reasoning.
    \item \textbf{Material Selection:} The agent currently selects from a predefined subset of common optical glasses. Expanding this to full industrial catalogs (e.g., Schott or Ohara) or incorporating temperature-dependent refractive index data ($dn/dT$) would significantly increase the search space complexity.
    \item \textbf{Computational Latency:} The system's performance is partially bottlenecked by the API call latency of commercial ray-tracing kernels. Developing a lightweight, differentiable proxy for initial structural screening could further accelerate the self-evolving process.
\end{itemize}

\subsection{Broader Impact}

\subsubsection{Intended Uses}
\textsc{LensDesigner} aims to democratize high-end optical engineering through several key applications:
\begin{itemize}
    \item \textbf{Accelerating R\&D:} Reducing the design cycle for consumer electronics, medical imaging, and automotive sensors from weeks to minutes.
    \item \textbf{Educational Support:} Serving as an interactive pedagogical tool for students to explore the trade-offs between lens complexity and imaging performance.
    \item \textbf{Cross-Disciplinary Innovation:} Enabling researchers in fields such as robotics or biology to customize specialized imaging systems without requiring years of domain-specific training.
\end{itemize}

\subsubsection{Potential Risks and Mitigations}
The automation of physical design carries inherent responsibilities:
\begin{itemize}
    \item \textbf{Dual-Use Concerns:} Advanced engineering tools could theoretically be applied to develop components for surveillance or specialized weaponry. We advocate for integrating safety alignment protocols within the foundation models to flag requests that violate security norms.
    \item \textbf{Over-reliance on Simulation:} AI-generated designs may not fully account for manufacturing tolerances (e.g., centering errors, mounting stresses). We emphasize that \textsc{LensDesigner} is an \textit{augmentative tool}; final designs must undergo rigorous optomechanical tolerance analysis by experts before fabrication.
    \item \textbf{Environmental Impact:} Large-scale agentic frameworks consume significant computational power. Our self-evolving mechanism mitigates this by distilling experience into a compact library, reducing the need for brute-force optimization in future tasks.
\end{itemize}



\end{document}

%% file: color_box_define.tex
\newtcolorbox{observationbox}[1][]{
        colback=envfill,
        colbacktitle=envfill,
        colframe=envborder,
        arc=5pt,
        fontupper=\small,
        fonttitle=\bfseries\color{black},
        boxrule=0.5mm,
        boxsep=1mm,
        width=\linewidth,
        breakable,
        title={Observation \hfill #1},
        rounded corners,
        toptitle=0.7mm,
        bottomtitle=0.7mm
}
\newtcolorbox{goldpatchbox}[1][]{
        colback=goldpatchfill,
        colbacktitle=goldpatchfill,
        colframe=goldpatchborder,
        arc=5pt,
        fontupper=\small,
        fonttitle=\bfseries\color{black},
        boxrule=0.5mm,
        boxsep=1mm,
        width=\linewidth,
        breakable,
        title={Gold Patch \hfill #1},
        rounded corners,
        toptitle=0.7mm,
        bottomtitle=0.7mm
}
\newtcolorbox{issuebox}[1][]{
        colback=issuefill,
        colbacktitle=issuefill,
        colframe=issueborder,
        arc=5pt,
        fontupper=\small,
        fonttitle=\bfseries\color{black},
        boxrule=0.5mm,
        boxsep=1mm,
        width=\linewidth,
        breakable,
        title={Issue \hfill #1},
        rounded corners,
        toptitle=1mm
}
\newtcolorbox{agentbox}[1][]{
        colback=agentfill,
        colbacktitle=agentfill,
        colframe=agentborder,
        arc=5pt,
        fontupper=\small,
        fonttitle=\bfseries\color{black},
        boxrule=0.5mm,
        boxsep=1mm,
        width=\linewidth,
        breakable,
        title={SWE-agent \hfill #1},
        rounded corners,
        toptitle=1mm,
        lower separated=false
}
\newtcolorbox{fileviewerbox}[1]{
        enhanced,
        breakable,
        boxrule = 1.5pt,
        fontupper = \small,
        fonttitle = \bf\color{black},
        arc = 5pt,
        rounded corners,
        colframe = black,
        colbacktitle = swecream,
        colback = swecream,
        title = #1,
        left=4pt 
}
\newtcolorbox{promptbox}[1]{
    enhanced,
    breakable,
    boxrule=1pt,  
    fontupper=\small,
    fonttitle=\bfseries\color{black},
    arc=3pt,  
    rounded corners,
    colframe=black,
    colbacktitle=swecream,
    colback=swecream,
    title=#1,
    left=2mm,  
    right=2mm,  
    top=1mm,  
    bottom=1mm  
}

%% file: supp/benchmark_examples.tex
{
\captionsetup{type=figure}
\begin{fileviewerbox}{Linting Error Message}
\begin{Verbatim}[breaklines=true]
{
  "easy": {
    "data": [
      {
        "id": 1,
        "fov": 30,
        "fnum": 4.5,
        "y": 10.8,
        "effl": 40.31,
        "rms_target": 0.048,
        "distortion_pct": 5,
        "difficulty": "easy",
        "design": "Please design an F/4.5 lens with a FOV of 30 degrees, half image height y=10.8mm, target RMS < 0.048mm, and distortion \leq 5%."
      },
      {
        "id": 11,
        "fov": 25,
        "fnum": 5.9,
        "y": 2.0,
        "effl": 9.02,
        "rms_target": 0.056,
        "distortion_pct": 5,
        "difficulty": "easy",
        "design": "Could you design a lens for me with FOV=25 degrees and F/5.9, featuring a half image height y=2.0mm, target RMS < 0.056mm, and distortion \leq 5%?"
      }
    ]
  },

...

  "medium": {
    "data": [
      {
        "id": 16,
        "fov": 70,
        "fnum": 2.6,
        "y": 5.3,
        "effl": 7.57,
        "rms_target": 0.057,
        "distortion_pct": 5,
        "difficulty": "medium",
        "design": "I need you to design a lens with a 70-degree FOV and F/2.6. The half image height should be y=5.3mm, target RMS < 0.057mm, and distortion \leq 5%."
      },
      {
        "id": 17,
        "fov": 105,
        "fnum": 4.5,
        "y": 3.0,
        "effl": 2.3,
        "rms_target": 0.03,
        "distortion_pct": 5,
        "difficulty": "medium",
        "design": "Design an F/4.5 lens with a 105-degree FOV for me, where half image height y=3.0mm, target RMS < 0.03mm, and distortion \leq 5%."
      }
    ]
  },

...

  "hard": {
    "data": [
      {
        "id": 10,
        "fov": 110,
        "fnum": 1.9,
        "y": 2.0,
        "effl": 1.4,
        "rms_target": 0.048,
        "distortion_pct": 5,
        "difficulty": "hard",
        "design": "Can you design a lens with FOV=110 degrees, F/1.9, half image height y=2.0mm, target RMS < 0.048mm, and distortion \leq 5/%?"
      },
      {
        "id": 13,
        "fov": 10,
        "fnum": 2.0,
        "y": 3.6,
        "effl": 41.15,
        "rms_target": 0.038,
        "distortion_pct": 5,
        "difficulty": "hard",
        "design": "I would like you to design a lens featuring a 10-degree FOV and F/2.0, with a half image height y=3.6mm, target RMS < 0.038mm, and distortion \leq 5%."
      }
    ]
  }
}

\end{Verbatim}
\end{fileviewerbox}
\captionof{figure}{
Some examples from our LensAreana benchmark. Queries are divided into three difficulty levels. Most challenging queries (e.g. ultra-wide-angle lenses and large-aperture lenses) are included. For each query, there is an RMS target to determine whether the result is successful or not.
}
\label{fig:benchmark_examples}
}